\documentclass{article}
\usepackage{arxiv}

\usepackage{listings}

\usepackage[utf8]{inputenc} 
\usepackage[T1]{fontenc}    
\usepackage{hyperref}       
\usepackage{url}            
\usepackage{bm}            
\usepackage{mathtools}
\usepackage{booktabs}       
\usepackage{amsfonts}       
\usepackage{microtype}      
\usepackage{lipsum}
\usepackage{graphicx}
\graphicspath{ {./images/} }
\usepackage[LGR,T1]{fontenc}
\usepackage[utf8]{inputenc}   
\usepackage{xcolor}

\newcommand{\textgreek}[1]{\begingroup\fontencoding{LGR}\selectfont#1\endgroup}

\title{How Many Words Does ChatGPT Know? \\   
The Answer is ChatWords} 

\author{
 Gonzalo Mart\'inez \\
  Universidad Carlos III de Madrid\\
  28911 Madrid, Spain \\
  \texttt{gonzmart@pa.uc3m.es} \\
  \And  
 Javier Conde \\
  Universidad Polit\'ecnica de Madrid\\
  28040 Madrid, Spain \\
  \texttt{javier.conde.diaz@upm.es}
  \And
  Pedro Reviriego \\
  Universidad Polit\'ecnica de Madrid\\
  28040 Madrid, Spain \\
  \texttt{pedro.reviriego@upm.es} \\
  \And
  Elena Merino-G\'omez\\
  Universidad de Valladolid \\
  47011 Valladolid, Spain \\  
  \texttt{elena.merino.gomez@uva.es}
  \And
  Jos\'e Alberto Hern\'andez\\
  Universidad Carlos III de Madrid\\
  28911 Madrid, Spain \\
  \texttt{jahgutie@it.uc3m.es} \\
  \And
  Fabrizio Lombardi \\
  Northeastern University\\
  02115 Boston, US \\
  \texttt{lombardi@ece.neu.edu} \\
}

\begin{document}
\maketitle
\begin{abstract}

The introduction of ChatGPT has put Artificial Intelligence (AI)  Natural Language Processing (NLP) in the spotlight. ChatGPT adoption has been exponential with millions of users experimenting with it in a myriad of tasks and application domains with impressive results. However, ChatGPT has limitations and suffers hallucinations, for example producing answers that look plausible but they are completely wrong. Evaluating the performance of ChatGPT and similar AI tools is a complex issue that is being explored from different perspectives. \\
In this work, we contribute to those efforts with ChatWords, an automated test system, to evaluate ChatGPT knowledge of an arbitrary set of words. ChatWords is designed to be extensible, easy to use, and adaptable to evaluate also other NLP AI tools. ChatWords is publicly available and its main goal is to facilitate research on the lexical knowledge of AI tools. The benefits of ChatWords are illustrated with two case studies: evaluating the knowledge that ChatGPT has of the Spanish lexicon (taken from the official dictionary of the "Real Academia Espa\~{n}ola") and of the words that appear in the Quixote, the well-known novel written by Miguel de Cervantes. The results show that ChatGPT is only able to recognize approximately 80\% of the words in the dictionary and 90\% of the words in the Quixote, in some cases with an incorrect meaning. The implications of the lexical knowledge of NLP AI tools and potential applications of ChatWords are also discussed providing directions for further work on the study of the lexical knowledge of AI tools. 

\end{abstract}


\section{Introduction}
\label{sec:introduction}

The introduction of generative AI tools capable of creating images from text descriptions such as DALL-E or Stable Diffusion \cite{SurveyDiffusionModels} and AI natural language processing tools based on large language models such as ChatGPT have put generative AI in the spotlight. ChatGPT reached one million users in the first week after its introduction and more than 100 million within its first two months, thus becoming the computing technology with the fastest adoption ever \cite{ChatGPT_IS1}. ChatGPT is now being used in a myriad of applications, and the technology has been incorporated into a variety of products \cite{ChatGPTOverview2}. 

ChatGPT achieves impressive performance and can answer a wide variety of questions on many different subjects as well as keep sophisticated conversations, translate or summarize text, follow instructions, or generate code among many other tasks \cite{ChatGPTOverview}. However, it also has limitations and suffers from hallucinations, so producing wrong responses that contain incorrect or false data but they look plausible \cite{ChatGPTFailures,Hernandez2023}. Therefore, significant efforts are being made to better understand its capabilities and limitations \cite{ChatGPTFailures}. Recent studies try to automate most of the testing to be able to evaluate ChatGPT performance over thousands or tens of thousands of samples from different publicly available datasets, for example in different types of questions and answer tasks \cite{ChatGPTEvaluation1,ChatGPTEvaluation2}. Automation enables not only testing large datasets but also evaluating different ChatGPT parameters and contexts. These studies focus on the performance of specific tasks such as question and answer and instructions. 

For AI tools that generate content such as ChatGPT, it is also relevant to understand which type of content is generated and in particular its fidelity and diversity \cite{FideltyDiversity}, especially when the AI tools are used massively, and their content populates the Internet \cite{martínez2023combining}. Moreover, the use of AI-generated data for training can lead to performance degradation and even to a model collapse \cite{shumailov2023curse,martínez2023understanding}. In the case of ChatGPT, analyzing the type of content generated is complex because it depends on the task, context, and language \cite{HCC1}. This has triggered research, for example, to try to understand phonological aspects \cite{HCC2}, linguistic patterns \cite{HCC3} or lexical richness \cite{reviriego2023playing} of AI-generated text and comparing it with those of human-written text.    

In this article, we contribute to the evaluation of ChatGPT assessing its knowledge of the words of a language. One of the basic features of a language is its vocabulary (words). The Real Academia Espa\~{n}ola (RAE) recognizes more than 90,000 different words\footnote{\url{https://www.rae.es/obras-academicas/diccionarios/diccionario-de-la-lengua-espanola}}, the Oxford English Dictionary recognizes 171,476 words\footnote{\url{https://languages.oup.com/}} in use and the Larousse Illustr\'e 63,800 words\footnote{\url{https://www.editions-larousse.fr/livre/grand-larousse-illustre-2023-9782035938718}} in French. However, the average person uses only a small fraction of those \cite{VocabSize1,VocabSizeSpanish}, typically between one-third and one-half for the most common Indoeuropean languages. Books can also be categorized or ranked based on the richness of its vocabulary. For instance, the Spanish Quixote \cite{Quijote} (Don Quijote de la Mancha) comprises 382,477, words, of which 22,632 are different words, so reading Quixote is a complex task.

Therefore, it is of interest to evaluate the knowledge that AI tools in general, and ChatGPT in particular have of the words of languages. Interestingly, this process can be automated as the list of words is typically available in dictionaries and prompts can be automatically designed such that ChatGPT produces answers that can be analyzed by a simple parser. In this work, we present ChatWords a tool to automatically test the knowledge that ChatGPT has of an arbitrary set of words. ChatWords has been designed with a modular approach to enable customizations and additions and to facilitate the evaluation of different lexicons and AI tools. The potential applications include assessing the knowledge of the lexicon of different languages, but also of those of specific domains, or even of incorrect words that may have been wrongly learned by ChatGPT. ChatWords is open-source and publicly available\footnote{\url{https://github.com/WordsGPT/ChatWords}}. To illustrate the benefits of ChatWords, it has been used to evaluate the knowledge that ChatGPT has of the Spanish lexicon and of the words in \textit{The Quixote} have been evaluated. The results show that ChatGPT does not recognize approximately 20\% of the Spanish lexicon. 

The rest of the paper is organized as follows. Section \ref{sec:Lexicon} briefly introduces the importance of words in a language and how they are used. Section \ref{sec:Testing} discusses the testing of the knowledge that ChatGPT has of a word as well as the process that can be automated together with the limitations that automation introduces. ChatWords is presented in section \ref{sec:LexiChat} and its potential benefits are illustrated in section \ref{sec:Casestudies} with two case studies. The potential implications of the lexical knowledge of AI tools like ChatGPT are discussed in that section, followed by a brief analysis of applications for LexiChat in section \ref{sec:Applications}. The paper ends with the conclusion in section \ref{sec:conclusions}.


\section{Words: the bricks of a language} 
\label{sec:Lexicon}

"In the beginning was the Word ..." 

this statement from the Gospel of John in the New Testament is used to introduce the importance of words in languages such that our understanding and analysis of languages start from words \cite{singleton2016language}. The lexicon of any language, defined as the complete list of all its words, forms the very foundation of human language. It is derived from the Greek term "lexis" (\textgreek{λέξις}), which means "speech", "way of speaking" and "word". Words serve as the medium through which the meanings of universal knowledge are represented, making them an essential part of our heritage. For instance, words classified as obsolete, often excluded from general dictionaries of “current language,” are crucial for understanding texts from different eras. The permanent loss of words can result in the loss of essential keys to understanding the world. Throughout history, efforts have been made to document and preserve every word of a language, including those that have fallen out of use, those currently in use, and new entries.

Since the 19th century, there has been a proliferation of dictionaries that collect historical lexicography in various languages. The desire for completeness has, at times, resulted in monumental works such as the TLL (Thesaurus Linguae Latinae)\footnote{\url{https://thesaurus.badw.de/tll-digital/tll-open-access.html}}, which was started in 1894 and is expected to be completed around 2050. However, attempts to carry out parallel work for the Greek TLG (Thesaurus Linguae Graecae) have encountered insurmountable difficulties \cite{GreekThesaurus1,GreekThesaurus2}. During the 20th century, the utilization of \textit{corpora} became increasingly prevalent. These compilations, encompassing a diverse array of language samples, serve various objectives, including the comprehensive documentation of languages' lexicons. This documentation not only offers insights from a historical standpoint, but it also provides valuable information about usage patterns. The emergence of digital technology and advancements in database management in the 21st century have facilitated a surge in the volume of entries and their systematic organization. This development has revitalized and propelled forward the \textit{corpora} projects that were originally initiated in the 20th century. Thus, the Oxford English Corpus (OEC) contains in its latest version nearly 2,1 billion words\footnote{\url{https://www.sketchengine.eu/oxford-english-corpus}}, the \textit{Corpus de Referencia del Espa\~{n}ol Actual} (CREA), contains more than 160 million forms\footnote{\url{https://www.rae.es/banco-de-datos/crea}} and the \textit{Corpus Diacrónico del Espa\~{n}ol} (CORDE) contains 250 million records\footnote{\url{https://www.rae.es/banco-de-datos/corde}} \cite{CREA}, the \textit{Digitales Wörterbuch der Deutschen Sprache (DWDS)} includes around 50 billion words from historical and contemporary collections\footnote{\url{https://www.dwds.de/r}}. However, in the case of the Corpus de \textit{r\'ef\'erence du fran\c{c}ais contemporain} (CRFC), a delay in its preparation is recognized in relation to other main languages (major languages) \cite{CRFC}.

Nowadays, online dictionaries allow for the inclusion of a vast repository of words that were previously declassified and removed from current use dictionaries of all languages due to the limitations of physical formats. While the deletion of words that had fallen out of use and the inclusion of new entries has been a constant throughout the history of dictionaries \cite{Palabrasaumentosupresion}, the general trend now is to keep all lemmas to the point of hinting “guarantee of comprehensiveness” \cite{mugglestone2011dictionaries}. Thanks to the possibilities of digital storage and their easy management, lexical repositories are now updated and categorized in real-time, allowing for the almost utopian goal of total conservation from now on.

This scenario is contrasted, however, with a possible impoverishment of language in use due to AI learning mechanisms. The overuse of certain words by humans, at the expense of others, could lead to biased learning of a language’s lexicon by AI, despite the language being inherently broader and more varied. The constant repetition and degradation processes evident in interactions with AI could compel it, in its pursuit of emulating natural human discourse, to use fewer and fewer words, despite its theoretical access to expansive online digital repositories. The ramifications associated with the potential loss of lexicon for universal knowledge, along with the imperative to safeguard each and every component of it, are succinctly encapsulated in Wittgenstein's renowned statement 5.6 from his \textit{Tractatus}: "The limits of my language mean the limits of my world." \cite{Wittgenstein}.

An interesting observation is that words are used with very different frequencies, and only a small fraction of the lexicon concentrates most of the utterances \cite{Zipf+2013}, following the so-called Zipf distribution. The Zipf distribution is the discrete version of the continuous Pareto distribution, used by the economist Pareto to describe the unfair distribution of wealth among people. The same Zipf behaviour is observed in multiple nature phenomena, including the Internet and social networks \cite{WordLex}. As discussed in the introduction, most persons only know a fraction of the words \cite{VocabSize1}. For Spanish, the number of words that an average person can recognize has been estimated to be around 30,000 which is approximately one-third of the words in the dictionary \cite{VocabSizeSpanish}.

\section{Testing if ChatGPT knows a word: automation and limitations} 
\label{sec:Testing}

To evaluate if ChatGPT knows a given word, for example, "dog" we can simply create an input prompt like, "Do you know the meaning of the word dog in English?" and let ChatGPT answer. Then read the answer to check if ChatGPT produces a valid description of the meaning. This, however, does not scale if we want to test all the words in the dictionary because we have to manually check ChatGPT’s response for each and every single word. Therefore, testing must be automated.

\subsection{Automating the testing process}

To automate testing, we need prompts that instruct ChatGPT to produce an answer that can be processed automatically. Therefore, the prompts must be designed such that responses can be easily parsed to extract the relevant information. In more detail, we can for example explicitly ask ChatGPT to answer only "yes" or "no" making our questions Boolean. Note that even Boolean questions may not be trivial and may require complex reasoning \cite{clark2019boolq}. This is not our case, we try to make questions as simple and direct as possible. For example, we use the following prompts:

\begin{enumerate}
    \item Prompt \#1: "Do you know the meaning of the word "X" in Spanish? Please answer yes or no."
    \item Prompt \#2: "Is "X" a correct word in Spanish? Please answer, yes or no."
    \item Prompt \#3: "Is "X" a valid word in Spanish? Please answer, yes or no."    
    \item Prompt \#4: "Is the word "X" in the Dictionary of the Real Academia Espa\~{n}ola (RAE)? Please answer yes or no."
\end{enumerate}

These prompts are simple, and the responses are easy to parse, so they can be processed automatically. More elaborate prompts can be used to ask for example for the meaning of the word and try to use metrics like perplexity \cite{Jelinek1977PerplexityaMO} to evaluate the confidence that the tool has in its response. Similarly, the response can be compared with the meaning of the word extracted from a dictionary. However, the automation of such checking process becomes significantly more complex. Therefore, we will use simple prompts like the ones listed above, leaving the use of more complex textual responses for future work. An intermediate approach is to get the meaning of the word from a dictionary and then ask ChatGPT to give a word with that meaning that starts with the first letters of the word. This may allow automation by checking if the word is part of ChatGPT's answer, so just like playing the alphabet game with ChatGPT. This could be the next step toward the automation of more elaborate prompts.

\subsection{Limitations of automation}

\begin{figure}[h]   
    \centering
    \includegraphics[width = 0.75\textwidth]{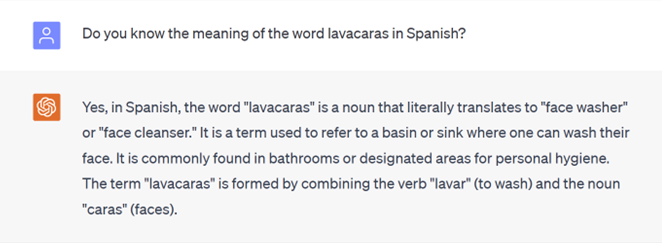}
    \caption{\textbf{Example of ChatGPT answering with an incorrect meaning for a word. "Lavacaras" means "flatterer" and ChatGPT answers with a meaning that seems taken from "Lava" (wash) and "Cara" (face). }}
    \label{Fig:Lavacaras}
\end{figure}

To understand the limitations introduced by  these simple prompts in the analysis, 100 words have been randomly selected from the dictionary of the Real Academia Espa\~{n}ola and they have been manually tested using additional prompts to check if the meaning of a given word by ChatGPT is correct when it answers yes. In more detail, the prompt: "Do you know the meaning of the word X in Spanish?" was used. In five of the words, the response of ChatGPT was an incorrect meaning; an example is shown in Figure \ref{Fig:Lavacaras}. Therefore,  the results presented in our analysis are likely to overestimate the lexical knowledge of ChatGPT. However, the percentage of such mistakes was only 5\% in our random sample and thus, simple prompts can be used to get an initial estimate of the lexical knowledge of ChatGPT. In any case, automating the checking of the meaning of the words rather than using a simple yes/no response is of interest to detect the words that ChatGPT misinterprets.

\section{ChatWords: automating lexical knowledge evaluation in ChatGPT} 
\label{sec:LexiChat}

The basic idea behind our tool is that once the testing of a word has been automated the same process can be run for a given lexicon such that all words are tested and all responses are parsed and compiled. Therefore, experiments can be defined to evaluate any arbitrary set of words for a specific AI model and configuration and then run. With this goal, ChatWords has been developed with two types of users in mind: 1) programmers or data scientists and 2) users who generally do not have programming skills, for example linguists, and want to use the tool from a simple graphical interface. The goal is to support both advanced users who want to modify the tool or even code part of the tool and users who want an easy-to-use web-based application.

In the rest of the section, we first describe the overall architecture of the tool and then describe both versions of the tool. The source code of ChatWords is available on a github repository\footnote{\url{https://github.com/WordsGPT/ChatWords}} offered under the GNU General Public License v3.0 (GPL-3.0). 

\subsection{Tool architecture}

\begin{figure}[h]   
    \centering
    \includegraphics[width = 0.50\textwidth]{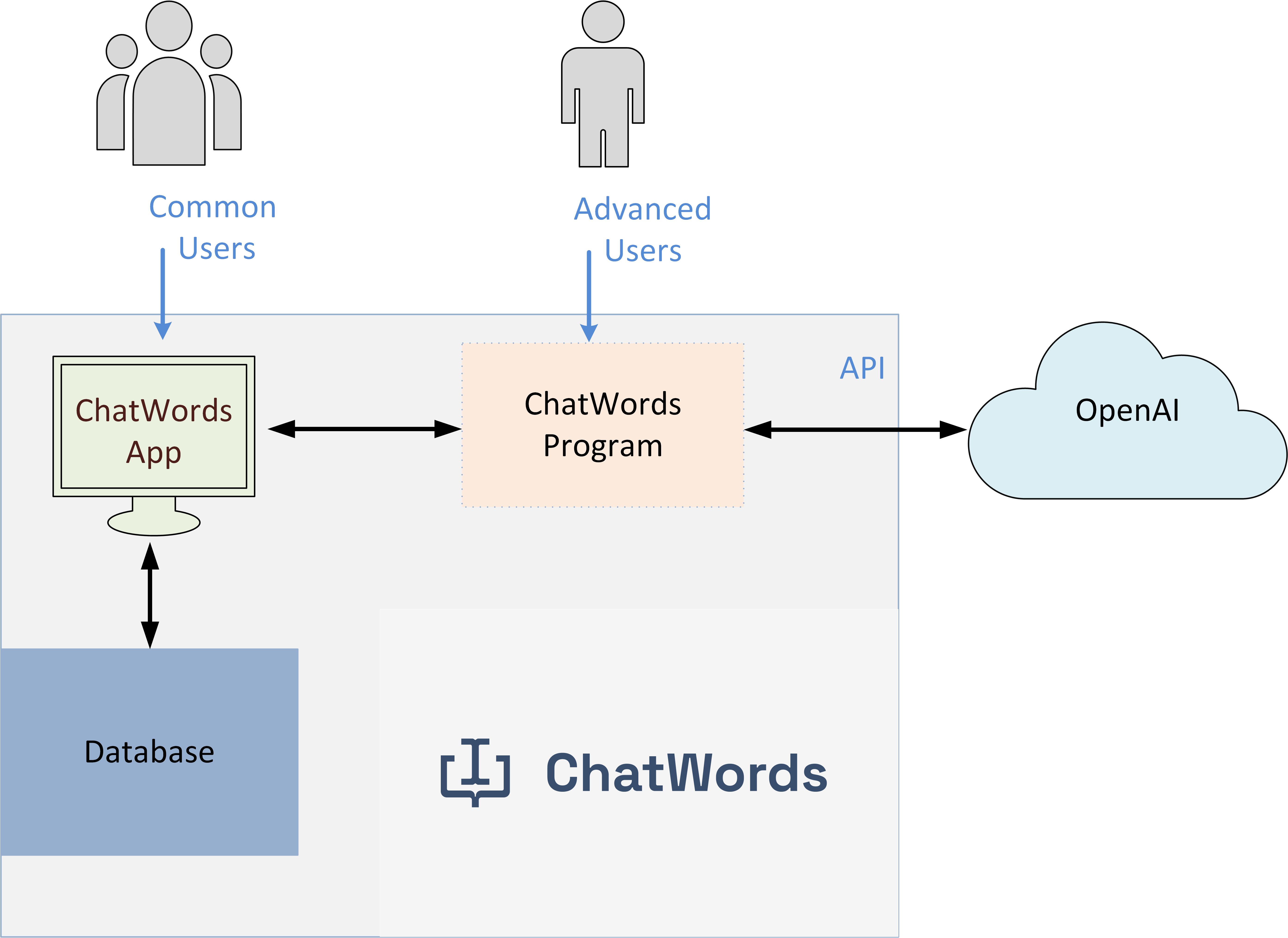}
    \caption{\textbf{Block diagram of ChatWords}}
    \label{Fig:Arch}
\end{figure}

The tool is divided into two parts, a program that interacts with ChatGPT using its API and a web application that interacts with the program to run experiments and store the results in a database. The interface between both parts is clearly defined, so that each part can be modified independently. The overall architecture is shown in Figure \ref{Fig:Arch}; users can interact with ChatWords through the ChatWords application to configure and run their experiments. This application calls a lower-level program that executes the experiments and sends the results to the Web application that then writes them in a simple database. The user can then access the results at any time from the database and keep a record of all experiments and results. Advanced users can directly run the lower-level program from a shell or command line and get the results in a file. 

The interfaces between the ChatWords program and the ChatWords application are clearly defined to enable programmers to create their own low-level programs; for example, advanced users may create programs that evaluate other AI tools instead of ChatGPT, or use a different programming language. Similarly, the low-level program developed in Python can be used with another implementation of the ChatWords application. This flexibility combined with the availability of the source code for both components is intended to facilitate extensions and modifications of ChatWords.

\subsection{ChatWords Program}

\begin{figure}  
    \centering
    \includegraphics[width = \textwidth]{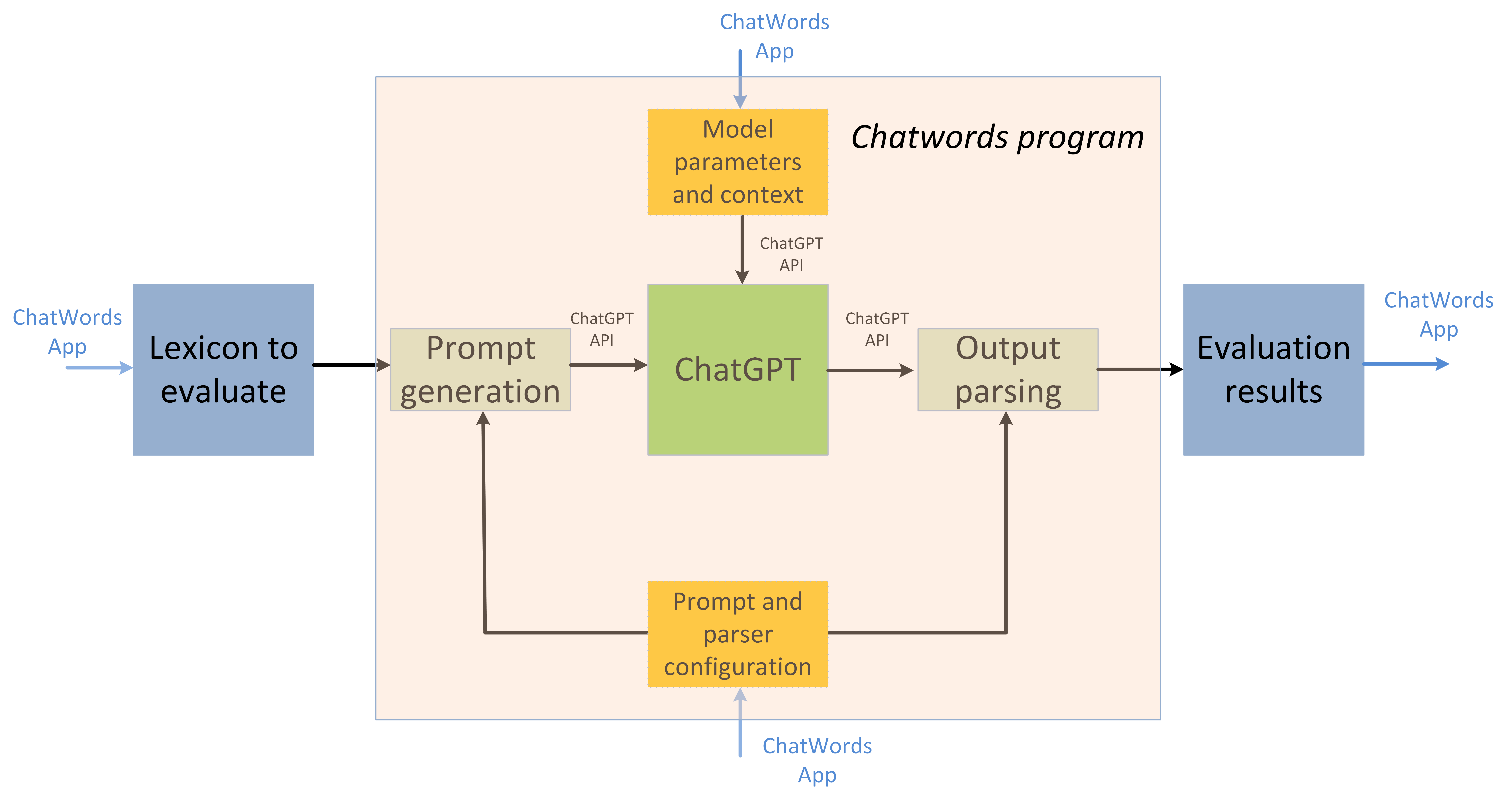}
    \caption{\textbf{Block diagram of the ChatWords program}}
    \label{Fig:LextChat}
\end{figure}

The block diagram of the program is shown in Figure~\ref{Fig:LextChat}. The parameters and lexicon to evaluate are provided by the ChatWords application and the results are sent back to the ChatWords application. For each word received, prompts are generated and sent to ChatGPT using OpenAI's API. The configuration of ChatGPT is also determined by the user with the ChatWords application. The architecture is modular and intended to provide flexibility to users who want to add new functionalities. In addition to the ChatWords application, the ChatWords program can also be run on its own from the command line.  

The program is written in Python and can be run from a shell or in other ways outside the web; this gives the user the flexibility to adapt it to their needs. Requests are processed asynchronously, reaching a maximum speed of 3,500 prompts processed per second, reaching the limits of OpenAI's API\footnote {OpenAI API limits: \url{https://platform.openai.com/docs/guides/rate-limits/overview}, last time accessed September 2023.}. Users can modify the speed of the experiment based on their needs or capabilities.

\subsection{ChatWords application}

One of the main design goals of ChatWords is that it can be used by people without programming skills; to that end, a web-based user interface is also provided. This version is programmed in Nest.js (a progressive Node.js framework for the development of web applications) for the back-end, and Angular (a web development JavaScript framework) for the front-end. The application stores the configuration and results for each experiment in a simple and easy-to-configure database; the application can be easily installed and run. The user has a web interface to select the configuration parameters, input file and to load results from previous experiments. 

The tool is designed to support additional configuration parameters as needed; the prompts are also configurable by the user or the programmer. It also provides a template to ease the creation of new experiments for programmers. It is agnostic to the model used for the experiments; even new models not developed yet can be integrated easily. It acts as an intermediary system that helps to create experiments, configure their parameters, load the words that want to be tested, save the results after they have been obtained, and present the results in real time through its web interface.

Chatwords aims to facilitate collaboration between linguists (often lacking technical programming skills) and data engineers. To this end, the ChatWords application allows linguists to set up experiments via a web interface and to upload large amounts of words. From the interface, the experiment can be started, stopped, and the results are shown in real time as they are generated; nevertheless, data engineers provide ChatWords programs with the logic of each experiment. ChatWords is not only model agnostic, but it is also agnostic to the configuration of the experiment itself. For example, in the ChatGPT API it is possible to select parameters such as temperature or limit the number of tokens of the answer. Therefore, the graphical interface of ChatWords adapts itself to each experiment starting from a JSON configuration file provided by the data engineer that includes metadata about the experiment, such as for example:

\begin{center}
\begin{tabular}{c}
\begin{lstlisting}[
    basicstyle=\scriptsize,
     linewidth=0.6\textwidth
]
{

    "name": "Template experiment",
    "description": "Description of template experiment",
    "configuration": {
        "model": {
            "type": "select",
            "options": [
                {"ChatGPT 3.5": "ChatGPT 3.5"},
                {"ChatGPT 4": "ChatGPT 4"}
            ] 
        },
        "temperature": {
            "type": "number",
            "name": "Configuration param 1",
            "placeholder": "Introduce the configuration parameter 1",
            "value": 0,
            "step": 0.1,
            "min": 0,
            "max": 1
        }
    }
}
\end{lstlisting}
\end{tabular}
\end{center}

\begin{figure}[h]   
    \centering
    \includegraphics[scale=0.2]{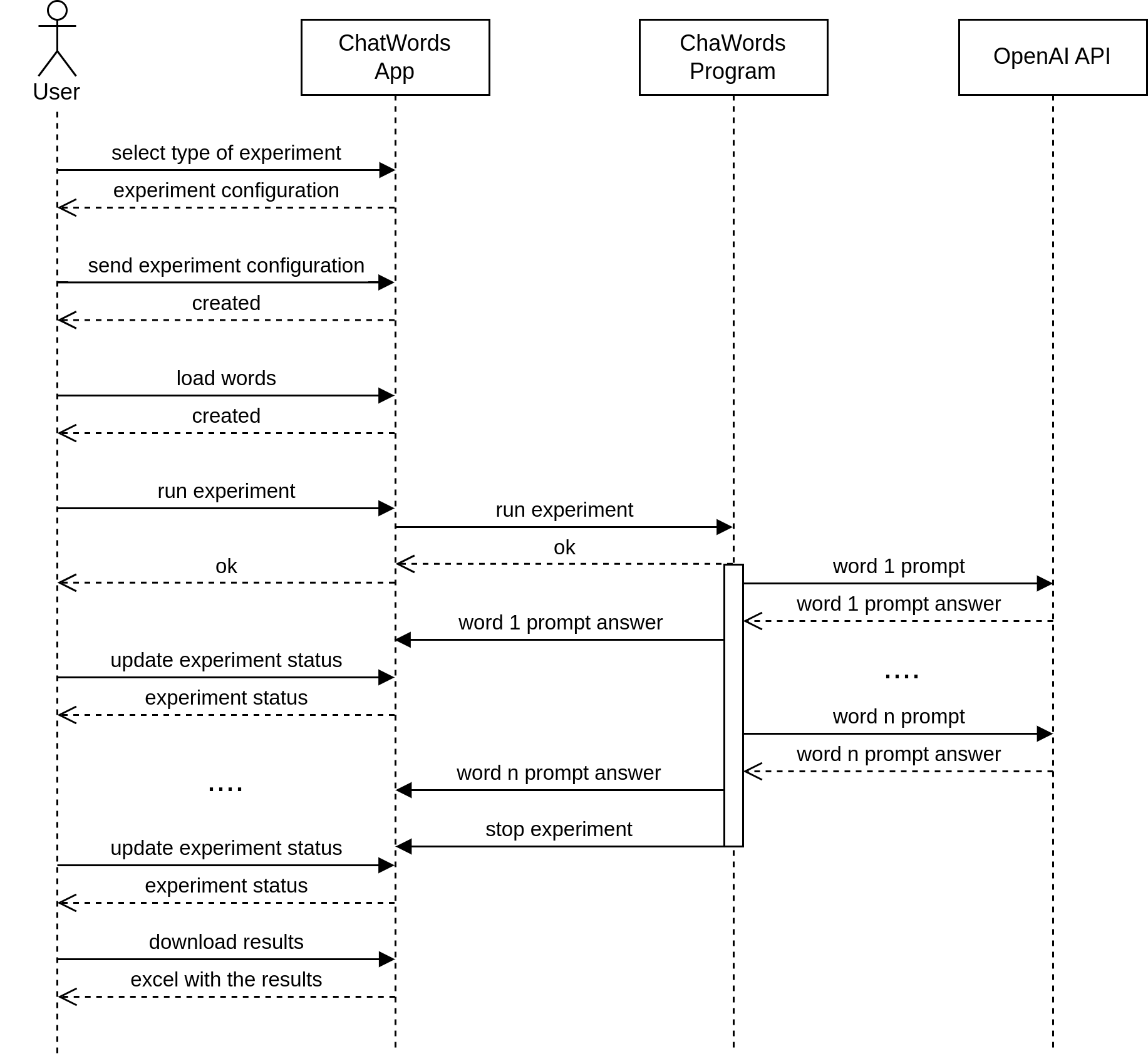}
    \caption{\textbf{Diagram for setting up and running an experiment}}
    \label{Fig:seq_diagram}
\end{figure}

The metadata includes the name of the experiment, the description, and the input of the form to be populated from the graphical interface. According to the previous template, ChatWords renders a form with a selector to choose the ChatGPT model (GPT 3.5 or GPT 4), and the temperature (controls the degree of randomness in the generated text.). Figure \ref{Fig:seq_diagram} shows a diagram with all the steps involved in setting up and running an experiment.

\section{Case studies: Spanish Lexicon and the Quixote}
\label{sec:Casestudies}

To validate the functionality of ChatWords and illustrate the benefits of lexicon knowledge evaluation, the Spanish language lexicon and the words used in \textit{The Quixote} have been used as case studies; so they are discussed in the following subsections.  

\subsection{Spanish Lexicon}

The most authoritative reference source in Spanish is the dictionary produced by the Real Academia de la Lengua Espa\~{n}ola (RAE). The Academy is a typical Enlightenment institution, founded in Madrid in 1713. It was inspired by the precepts of the Académie Française, founded in the previous century, to regulate and perfect the French language. However, since its inception, the interest of Spanish academics focused, among other goals, on the development of a dictionary, whose first edition was published in 1780 \cite{RAEHistoria}.

Twenty-three editions of the Dictionary of the Spanish Language (Diccionario de la lengua Espa\~{n}ola: DLE) have been published since the 18th century. The most recent one is the twenty-third, published in 2014. The dictionary includes lexical entries that belong to the entire Spanish-speaking context. The analysis of the lexicon and the decisions on modifications, additions, the inclusions of new meanings, amendments, etc. are adopted by consensus among the 23 academies of the Spanish language, present in Spain, America, the Philippines, and Equatorial Guinea. The successive updates of the dictionary are reflected through the different electronic versions. The current one is version 23.6, which includes new lemmas, completes etymologies, adds new meanings, or amends different aspects of previous versions\footnote{\url{https://dle.rae.es/docs/Novedades_DLE_23.6-Seleccion.pdf}}. The dictionary provides information, among other issues, on the specific geographical use of words that are not used generally in the Spanish-speaking community, as well as terms that are rarely used, archaic, out of use, poetic, vulgar, etc.

The words of the RAE dictionary were downloaded from the RAE website\footnote{\url{https://dle.rae.es/}} using a modified version of a publicly available scrapper\footnote{\url{https://github.com/JorgeDuenasLerin/diccionario-espanol-txt}} that was modified to remove the gender forms of words. The result is a list of 91,168 words that we use to evaluate ChatGPT. To achieve a more balanced version of the list, only the verbal infinitives are presented, eliminating all their inflectional variants, which in Spanish are approximately 80 forms per verb. Variants of gender and number have also been eliminated, using the singular masculine generic as unmarked lemmas. For example, the ‘abogado’ is the chosen lemma and its variants of gender, “abogada”, and number, “abogados”, “abogadas” are discarded. The final number of words included in the file used for evaluation is, in any case, very similar to that in the twenty-third edition of the dictionary, the most recent, with 93,111 entries \cite{DICRAE}.

\subsection{The Quixote}

In the second case study, we analyze a specific case, the words used in \textit{The Quixote}, the novel by Miguel de Cervantes. For this evaluation, the commemorative edition of the IV centenary of \textit{Don Quixote de la Mancha}, published by the RAE and the Association of Academies of the Spanish Language  \cite{Quijote}, has been used. This text showcases a spelling update while retaining archaisms, cultisms, vulgarisms, and certain outdated spellings. These have been retained when it was deemed that their removal would have resulted in misunderstandings or in diminished richness in textual interpretation. The book was processed extracting 22,632 unique words out of 382,477 words in the text.

\subsection{Evaluation}

Testing has been done with ChatGPT3.5turbo with no context and default settings for all the parameters except temperature that was set to zero for deterministic answers. The following prompts have been used:

\begin{enumerate}
    \item Prompt \#1: "Do you know the meaning of the word "X" in Spanish? Please answer yes or no."
    \item Prompt \#2: "Is "X" a correct word in Spanish? Please answer, yes or no."
    \item Prompt \#3: "Is "X" a valid word in Spanish? Please answer, yes or no."    
    \item Prompt \#4: "Is the word "X" in the Dictionary of the Real Academia Espa\~{n}ola (RAE)? Please answer yes or no."
\end{enumerate}

The first prompt is the most general one, while the fourth prompt is the most specific asking ChatGPT about RAE’s dictionary. Note also that the second and third prompts are very similar and should get the same response. They are used to check the consistency of the ChatGPT answers.

\begin{figure}[h]   
    \centering
    \includegraphics[scale= 0.4]{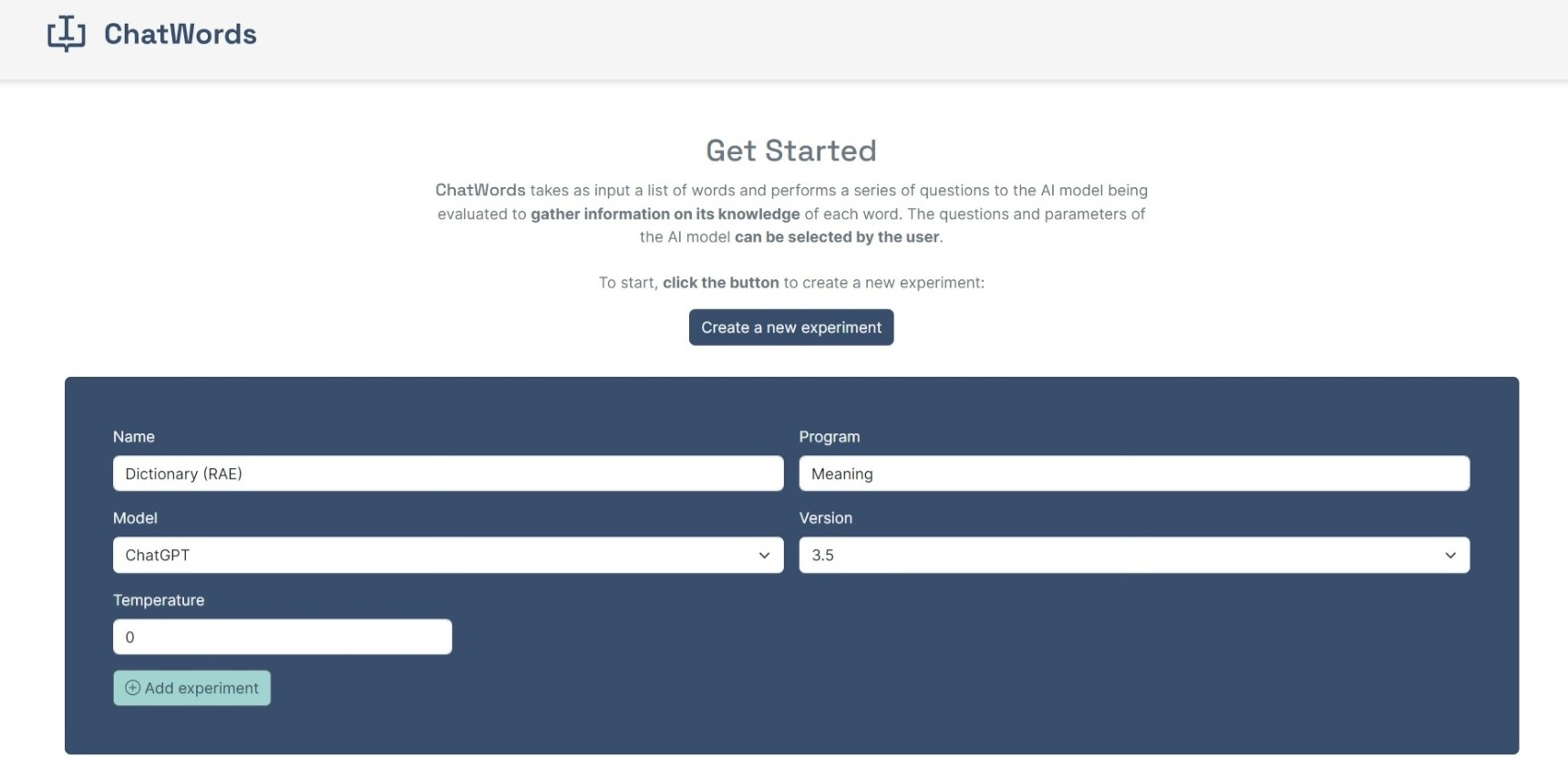} \\
    \includegraphics[scale= 0.24]{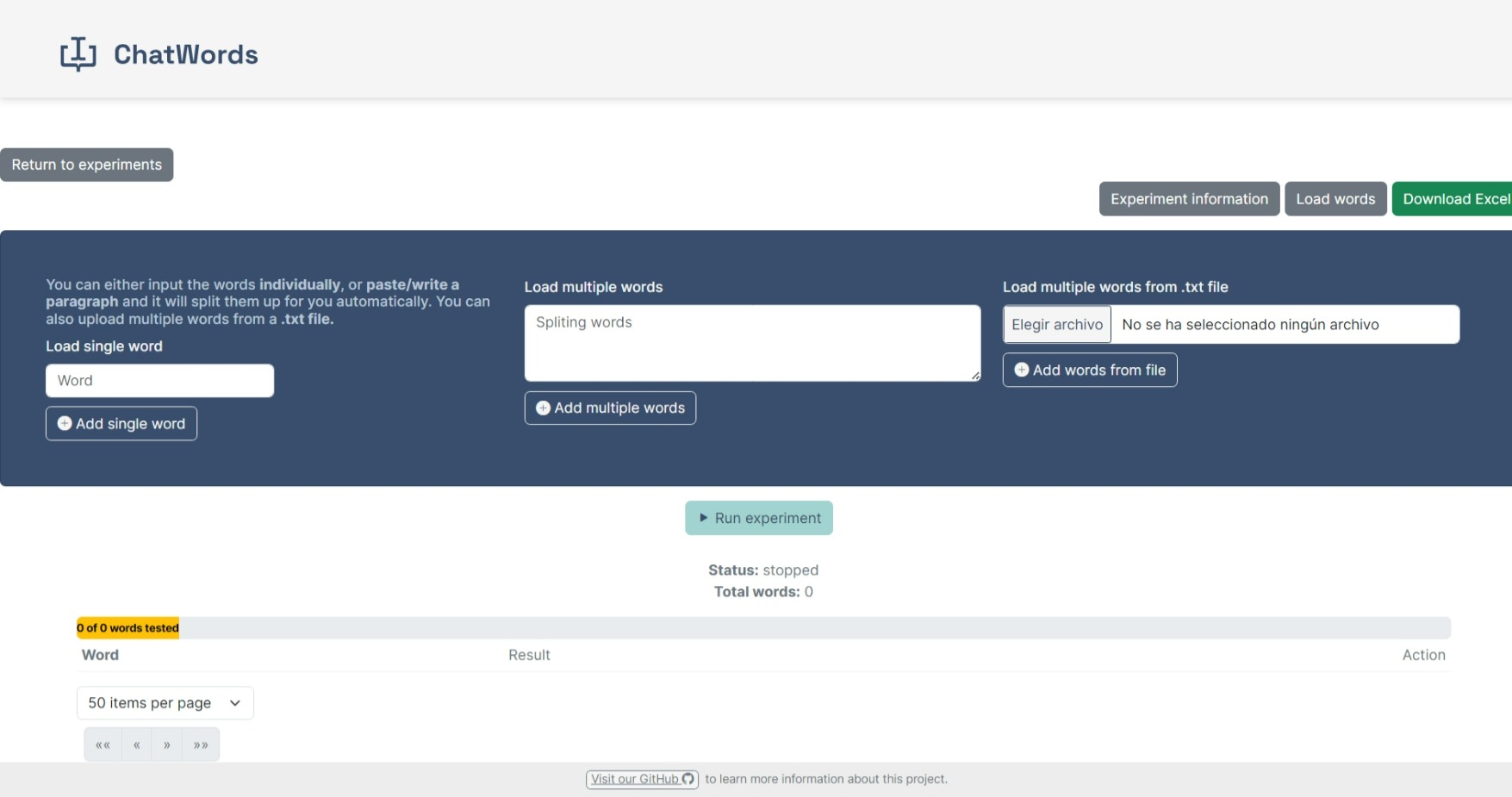}
    \caption{\textbf{Running a experiment in ChatWords: setting up the experiment (top) and loading the words and running the experiment (bottom)}}
    \label{Fig:screenshots}
\end{figure}

The Chatwords User Interfaces to generate the experiment are shown in Figure \ref{Fig:screenshots}. First the experiment is generated by selecting the AI model (ChatGPT), its parameters (version 3.5 and temperature = 0) and the low level program (meaning); then the words are loaded, in this case from a .txt file containing all words in the RAE’s dictionary.

\begin{figure}[h]   
    \centering
    \includegraphics[scale=0.55]{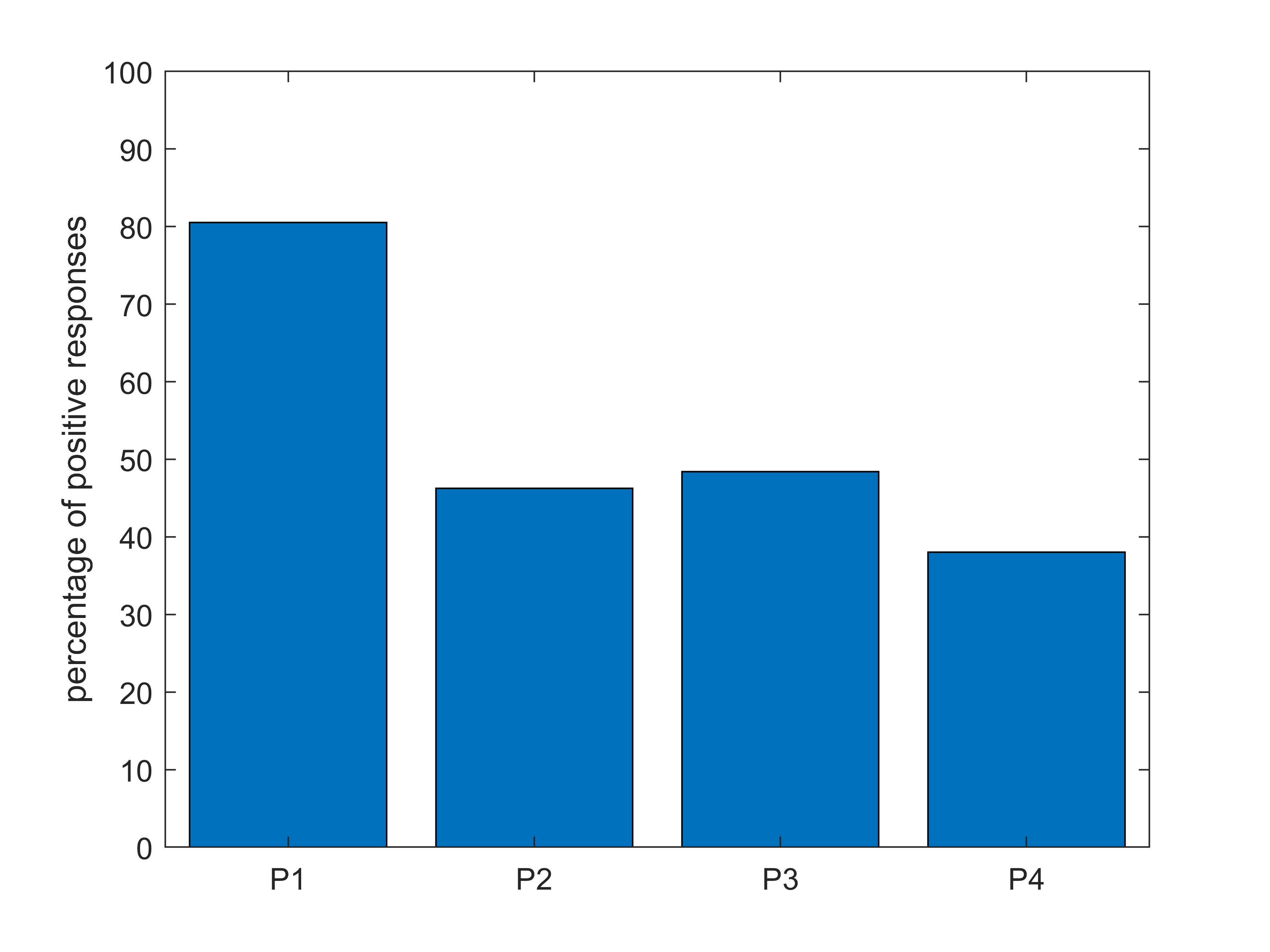}
    \includegraphics[scale=0.55]{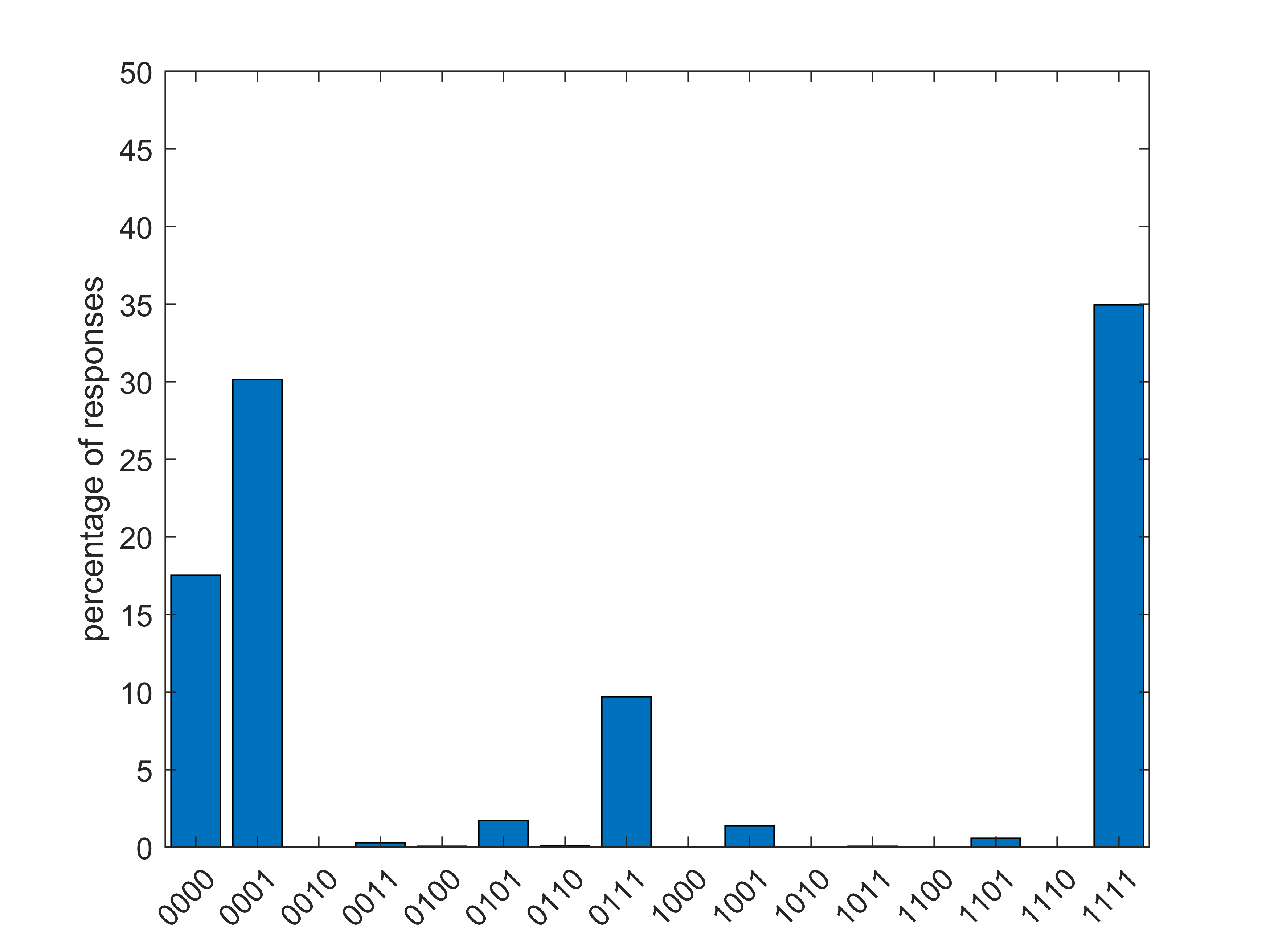}
    \caption{\textbf{Results for the words in the dictionary of the RAE: percentage of positive answers to each prompt (left), percentage of responses for each of the sixteen possible combinations of answers for the four prompts (right)}}
    \label{Fig:RAE}
\end{figure}

The results for the words in the dictionary are summarized in Figure \ref{Fig:RAE}. The first plot shows the percentage of positive responses to each of the prompts (P1 to P4). It can be observed that the first prompt has the largest fraction of positive responses while the last prompt the lowest value as expected; the second and third prompts are in the middle. Interestingly, the second and third prompts have slightly different percentages which show the variability of ChatGPT responses for similar questions. Even for the most generic prompt, there is a non-negligible fraction of negative responses, approximately 20\% which increases for the other prompts to 50-60\%. This shows that ChatGPT does not have a complete knowledge of the Spanish lexicon. The second plot shows the percentages for the sixteen possible combinations of the answers to the four prompts coded in binary as {P4,P3,P2,P1}. So, for example, '0001' corresponds to a positive answer to the first prompt and a negative for the other three. The fraction of words for which ChatGPT gives a positive response ('1111') for all four prompts drops to 35\%, mostly due to a large percentage of words not being identified as part of RAE's dictionary ('0xx1'). Combinations '0001' and '0111' have the largest values apart from '0000' which accounts for approximately 17.5\% of the words. This latest result confirms that ChatGPT does not seem to have any knowledge of a significant fraction of the words. It is also interesting to note that there is a small percentage of contradictory answers, for example, '0101' and '1101', in which ChatGPT says that a word is correct but not valid. Finally, it is important to note that as discussed in section \ref{sec:Testing}, automated testing is not fully accurate and for a small fraction of the words that ChatGPT recognizes it will have an incorrect meaning.

\begin{figure}[h]   
    \centering
    \includegraphics[scale=0.55]{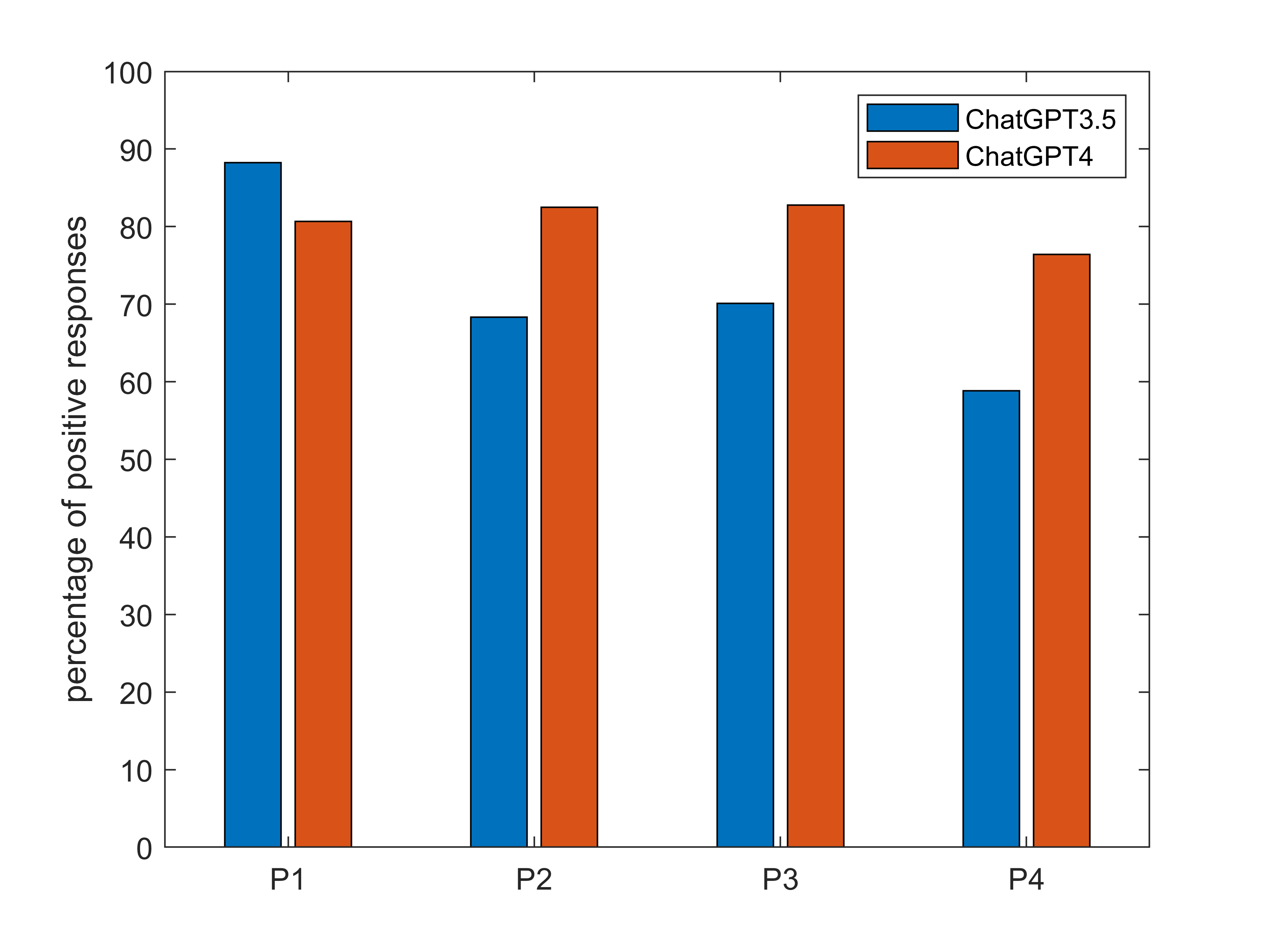}
    \includegraphics[scale=0.55]{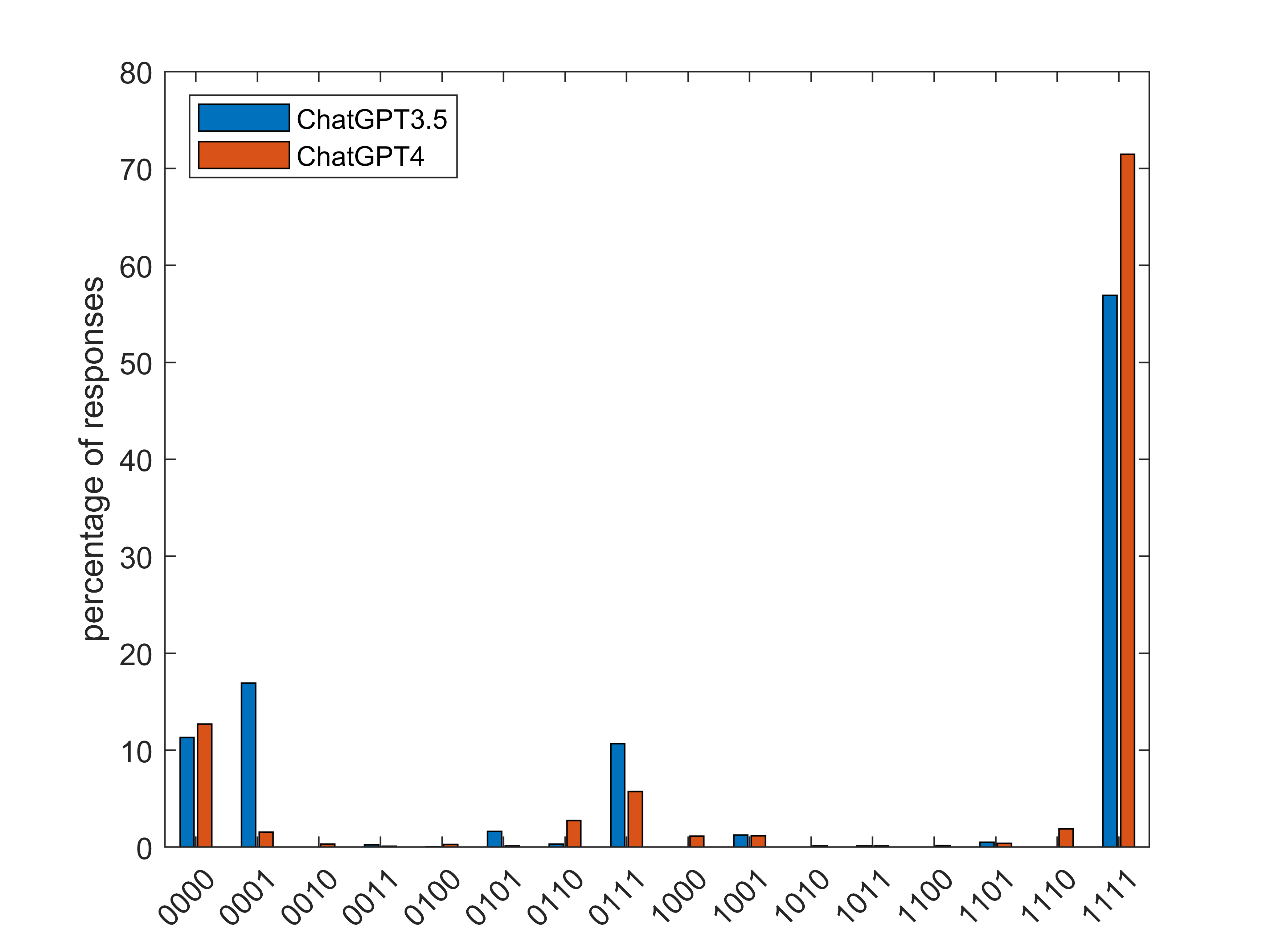}
    \caption{\textbf{Results for the words in \textit{The Quixote}: percentage of positive answers to each prompt (left), percentage of responses for each of the sixteen possible combinations of answers for the four prompts (right)}}
    \label{Fig:Quixote}
\end{figure}

For \textit{The Quixote}, the same testing was conducted but this time for both ChatGPT3.5turbo and ChatGPT4 to study whether the lexical knowledge has improved in the latest version of the tool. The results are shown in Figure \ref{Fig:Quixote}. For ChatGPT3.5turbo they follow similar trends as for the RAE dictionary: there is a significant number of negative answers that increase from the first to the last prompt; the most frequent combinations of the answers are also the same. The main difference is that the percentages of negative answers are lower than for the entire dictionary. This is reasonable because \textit{The Quixote} uses a fraction of the lexicon that includes most common words but only a small part of the less frequently used words, which are more prone to be unknown to ChatGPT. 

Focusing on the comparison between the two versions of ChatGPT, it can be observed that ChatGPT4 provides more consistent responses with fewer differences between the four prompts. The percentage of positives is lower for the first prompt, while it increases for the other three. These trends are more obvious on the right plot, in which the percentages of answers for which all prompts are all zeros or all ones increase from 11.3\% to 12.7\% and from 56.9\% to 71.4\%. The most notable reduction is for '0001' indicating that ChatGPT4 is more cautious to state that it knows the meaning of a word (P1). A random sample of 100 words was checked manually by asking ChatGPT4 for the meaning of the words and comparing it with that in the dictionary of the RAE. The number of false positives, so words for which ChatGPT4 states that it knows the meaning but then answers with an incorrect meaning, was reduced compared to ChatGPT3.5turbo. This is in line with the other results and confirms that ChatGPT4 provides more reliable answers. Overall it seems that lexical knowledge has not improved significantly from ChatGPT3.5turbo to ChatGPT4.

\subsection{Discussion and Implications}
\label{sec:Implications}

\begin{figure}[h]   
    \centering
    \includegraphics[scale=0.75]{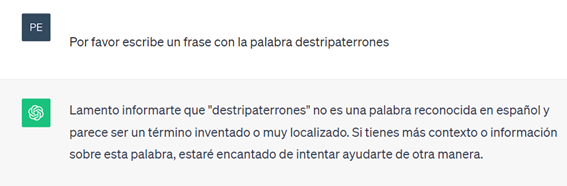}
    \caption{\textbf{Asking ChatGPT to write a sentence with a word it does not know "Destripaterrones" that is used in \textit{The Quixote} and included in the RAE's dictionary. ChatGPT refuses to write the sentence.}}
    \label{Fig:NotKown}
\end{figure}

The lexical knowledge of ChatGPT has potential implications for the Spanish language. It seems reasonable to assume that ChatGPT will not use words it does not know. To confirm this hypothesis, we ask ChatGPT to write a sentence using a word that it does not know and it refuses to do so, an example is shown in Figure \ref{Fig:NotKown}. Therefore, as ChatGPT and similar tools are increasingly used to write content that ends up on the Internet, future training datasets may have less lexical diversity. This may cause newer versions of ChatGPT to further reduce their lexicon leading to less rich content. There is already a strong debate on whether technology is reducing the lexicon used by newer generations. Although there are not sufficient studies to prove this hypothesis, the increasing speed of communication imposed by newer digital tools forces speakers to choose among their available lexicon \cite{LexicalAvailability} very quickly. This may introduce a bias towards using common words further increasing their frequencies, while other less common words that need more time and analysis, are thus less used increasing the gap between the available lexicon and the lexicon used. AI tools like ChatGPT may contribute to this trend by exposing users to a subset of words only, making the rest invisible and thus reinforcing the trend toward a reduction of the lexicon used, not only by newer AI tools but also by humans.

The mechanisms of linguistic evolution have produced modifications in phonetics, in its orthographic representation, as well as important semantic alterations of many words and expressions throughout the centuries. Semantic misunderstandings caused by the formal similarity of words or confusions caused by phonetic analogy or the inappropriate use of certain phrases that become obscure over time, are some of the phenomena that introduce new meanings, but which sometimes leads to a loss of precision in oral and written communication. Thus, the Latin expression “in flagranti [crimine]” is confused with the present participle, more common in Spanish, “fragante” [from lat. fragrans, fragrantis: that exhales an odor, a perfume]. The confusion results in the now accepted “in fraganti”, an adverbial phrase documented in Spanish at least since the 19th century\footnote{\url{https://www.rae.es/banco-de-datos/corde}}. The emergence of the Internet implies important changes in the speed of linguistic evolution, particularly in English, in its role as the lingua franca of the network, to the point of coining a new branch of linguistics: Internet Linguistics \cite{Alkadi2018EVOLUTIONOE}. Internet English, in some ways, could be compared to a typical koine since it is becoming the common language among users of native speakers of different languages, and, like all koines, tends to a general simplification, for the sake of a greater ability to spread and understand the message.


The consequences of the trend toward simplification are also evident in the AI preferences. In this regard, it is common for the interactions with AI carried out with the aim of improving the writing of a text, suggesting the use of more common or simpler words, sometimes resulting in changes or losses of nuances and meanings. Some examples of authentic Bing AI suggestions are:

\begin{itemize}
    \item Use “are made collectively” instead of “are made collegially” to use a more common word,
    \item Use “or modifies” instead of “or amends” to use a simpler word.
    \item Use “among other things” instead of “among other aspects” to use a more common expression.
\end{itemize}

Another interesting consideration is the relative knowledge of the lexicon across languages. Are the differences like to those observed in the performance for different tasks? For example, the performance for Spanish and English of ChatGPT-4 is very similar on some tasks\footnote{\url{https://openai.com/research/gpt-4}}, does the same apply to the lexicon? Or is there a stronger bias in the lexicon knowledge across languages? This is an interesting topic for future work.

\section{Lexical Analysis Applications}
\label{sec:Applications}

To illustrate the potential benefits of ChatWords this section discusses some applications in which lexical analysis can be useful. 

\subsection{Language analysis}
The most direct application of ChatWords is to evaluate the lexicon knowledge of ChatGPT across languages and how it depends on the version and configuration of ChatGPT. This would provide an understanding of how the tool captures the lexical richness of different languages. The results could also be compared with those of native speakers to put the results in context. 

More generally, instead of a simple yes/no answer, we could ask ChatGPT for the meaning of each word and then compare it against the meaning in a dictionary. This would provide a deeper understanding of the lexical knowledge of ChatGPT, not only if it knows the words but also their meanings. The automation of this application requires a more sophisticated processing of ChatGPT outputs that needs further study.

\subsection{Domain specific analysis}

Another potential use of ChatWords is to evaluate the knowledge of ChatGPT for the terminology and words used in specific domains such as Law, Medicine, Science, or Engineering in different languages. This can be of interest to users that rely on ChatGPT for specific areas of application.

\subsection{Names and places analysis}

\begin{figure}[h]   
    \centering
    \includegraphics[scale=0.7]{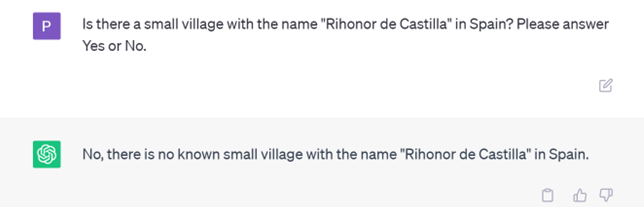}
    \caption{\textbf{Example of the evaluation of ChatGPT knowledge of places. Rihonor de Castilla is the Spanish part of a village that is half Spanish and half Portuguese (\url{https://en.wikipedia.org/wiki/Rihonor_de_Castilla}).}}
    \label{Fig:Rihonor}
\end{figure}

The tool can also be used to evaluate if ChatGPT knows the names of persons, places, or products. This can be useful for companies and for applications that use ChatGPT as these words that are not included in dictionaries. An example of the results obtained is shown in Figure \ref{Fig:Rihonor}, ChatGPT does not know a small village in the northwest of Spain.


\subsection{Slang and emerging lexicon analysis}

Another interesting aspect is to evaluate the knowledge that ChatGPT has of slang and new words being used by younger generations. For example, the lyrics of trendy trap or reggaeaton songs can be analyzed by ChatWords.


\subsection{Incorrect word analysis}

ChatWords is not only useful to assess the knowledge of the correct words; it can be used to check if ChatGPT has learned incorrect words that may be present in its training dataset. This can be checked by testing a set of false words \cite{ARC_nonwords, Wuggy} on ChatWords. It is of interest to understand the impact of using text that may not be correct during training. Although a detailed evaluation of the responses of ChatGPT to pseudowords that do not exist, is outside the scope of this work, we have used the Wuggy tool \cite{Wuggy} to generate a few pseudowords in Spanish and asked ChatGPT for their meanings. The answers seem to depend on the version of ChatGPT used and for some of them, the tool misinterpreted some words. For example, when asked about the word "seglo" the answer was "seglo means century in Spanish". Seglo does not exist in Spanish but it is quite similar to "siglo" which means century. Therefore, depending on the version and setting, it seems that ChatGPT can mistake invalid words for words that are similar and valid. The importance of accepting incorrect words as valid deserves further study and a detailed analysis of its implications.

\subsection{Impact of lexical knowledge on task performance}

A more subtle use of ChatWords would be to correlate ChatGPT performance in different tasks with its lexical knowledge. For example, for question and answer, summarization, or translation tasks, we could run ChatWords on the inputs to quantify the knowledge of the words used in the prompt and see if there is a correlation between lexical knowledge and task performance. A simple example of translation shows that the lack of lexical knowledge can lead to incorrect results and intuitively the same applies to other language processing tasks. 

\begin{figure}[h]   
    \centering
    \includegraphics[scale=0.7]{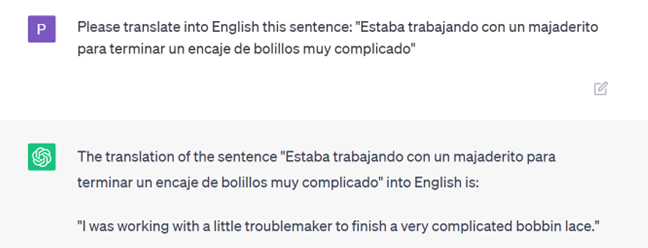}
    \caption{\textbf{Example of the impact of lack of lexical knowledge on other tasks. ChatGPT does not know the word "majaderito" that in Spanish is a type of bobbin lace and thus, it uses the translation of "majadero" which is completely unrelated.}}
    \vspace{-5mm}
    \label{Fig:Majaderito}
\end{figure}

\section{Conclusion}
\label{sec:conclusions}

In this paper, we have presented ChatWords, a tool to automate the lexical knowledge evaluation of ChatGPT. ChatWords is designed to be both user-friendly, requiring no programming skills, and highly flexible, allowing for easy modifications and extensions. Flexibility in terms of the prompts, AI tool configuration, result processing and integration with other AI tools is provided to facilitate research on the lexical knowledge of AI tools. The potential benefits of ChatWords have been demonstrated with two case studies: the evaluation of the Spanish lexicon and the words used in The Quixote. The applications of lexical knowledge evaluation of AI tools as well as its implications have also been discussed providing multiple directions for future work.

ChatWords is publicly available and we expect that its use will facilitate the evaluation of the lexical knowledge of AI tools in the applications discussed in the paper and beyond, as new horizons, investigations and scenarios will be proposed by other researchers

\section{Acknowledgements}
The authors would like to acknowledge the support of the FUN4DATE (PID2022-136684OB-C21/22) project funded by the Spanish Agencia Estatal de Investigacion (AEI) 10.13039/501100011033. The authors would like to thank Blanca Querol for the design of the ChatWords logo and graphical user interface.  

\bibliographystyle{unsrt}  


\bibliography{ChatGPTDic}

\end{document}